\def\year{2019}\relax
\documentclass[letterpaper]{article} 
\usepackage{aaai19}  
\usepackage{times}  
\usepackage{helvet}  
\usepackage{courier}  
\usepackage{url}  
\usepackage{graphicx}  

\usepackage{url}

\usepackage{amsmath}
\usepackage{algorithm}
\usepackage{algpseudocode}
\makeatletter
\def\BState{\State\hskip-\ALG@thistlm}
\usepackage{microtype}
\usepackage{graphicx}
\usepackage{subfigure}
\usepackage{booktabs} 
\usepackage{blindtext, xcolor}
\usepackage{comment}
\usepackage{natbib}
\usepackage{hyperref}

\usepackage[utf8]{inputenc} 
\usepackage[T1]{fontenc}    
\usepackage{hyperref}       
\usepackage{url}            
\usepackage{booktabs}       
\usepackage{amsfonts}       
\usepackage{nicefrac}       
\usepackage{microtype}

\begin{document}
%
\title{Scalable Recollections for Continual Lifelong Learning}
\author{Matthew Riemer, Tim Klinger, Djallel Bouneffouf, and Michele Franceschini\\
IBM Research\\
T.J. Watson Research Center, Yorktown Heights, NY\\
\{mdriemer, tklinger, djallel.bouneffouf, franceschini\}@us.ibm.com\\
} 
\maketitle
\begin{abstract}
Given the recent success of Deep Learning applied to a variety of single tasks, it is natural to consider more human-realistic settings. Perhaps the most difficult of these settings is that of continual lifelong learning, where the model must learn online over a continuous stream of non-stationary data. A successful continual lifelong learning system must have three key capabilities: it must \textit{learn and adapt} over time, it must \textit{not forget} what it has learned, and it must be \textit{efficient} in both training time and memory. Recent techniques have focused their efforts primarily on the  first two capabilities while questions of efficiency remain largely unexplored. In this paper, we consider the problem of efficient and effective storage of experiences over very large time-frames. In particular we consider the case where typical experiences are $O(n)$ bits and memories are limited to $O(k)$ bits for $k << n$.  We present a novel scalable architecture and training algorithm in this challenging domain and provide an extensive evaluation of its performance.  Our results show that we can achieve considerable gains on top of state-of-the-art methods such as GEM. 
\end{abstract}

\section{Introduction} 
A long-held dream of the AI community is to build a machine capable of operating autonomously for long periods or even indefinitely.  Such a machine must necessarily learn and adapt to a changing environment and, crucially, manage memories of what it has learned for the future tasks it will encounter. A spectrum of learning scenarios are available depending on problem requirements.  In lifelong learning \citep{LML} the machine is presented a sequence of tasks and must use knowledge learned from the previous tasks to perform better on the next. In the resource-constrained lifelong learning setting the machine is constrained to a small buffer of previous experiences.  Some approaches to lifelong learning assume that a task is a set of examples chosen from the same distribution \citep{PNN,Pathnet,DGR,LLGen,CMAML,DEN}. If instead the machine is given a sequence of examples without any batching, then this is called continual learning. In this paper we focus on this more challenging continual learning scenario. 

Continual learning \citep{Thrun94,Ring94,LML} has three main requirements: (1) continually adapt in a non-stationary environment, (2) retain memories which are useful, (3) manage compute and memory resources over a long period of time. Most neural network research has focused on methods to improve (1) and (2). In this paper we consider (3) as well and further investigate the role of efficient experience storage in avoiding the catastrophic forgetting \citep{CF} problem that makes (2) so challenging.  

Experience memory has been influential in many recent approaches.  For example, experience replay \citep{Lin92} was integral in helping to stabilize the training of Deep Q Learning on Atari games \citep{DQN}. Episodic storage mechanisms \citep{Prioritized,MFEC,NEC,iCaRL,GEM} were also some of the earliest solutions to the catastrophic forgetting problem in the supervised learning setting \citep{Murre92,Robins95}. 
Unlike approaches which simply focus on remembering representations of old tasks \citep{LwF,Riemer16,Kirkpatrick17}, episodic storage techniques achieve superior performance because of their ability to continually improve on old tasks over time as useful information is learned later \citep{GEM}.

All of these techniques try to use stored experiences to stabilize learning. However, they do not consider agents which must operate independently in the world for a long time. In this scenario, assuming the kind of high-dimensional data which make up human experiences, the efficient storage of experiences becomes an important factor. Storing full experiences in memory, as these methods do, causes storage costs to scale linearly with the number of experiences. To truly learn over a massive number of experiences in a non-stationary environment, the incremental cost of adding experiences to memory must be sub-linear in the number experiences.

In this paper we propose a scalable experience memory module which learns to adapt to a non-stationary \footnote{We assume the environment is non-stationary, but not adversarial. So past experiences can be helpful for learning future tasks.} environment and improve itself over time.  The memory module is implemented using a variational autoencoder which learns to compress high-dimensional experiences to a compact latent code for storage. This code can then be used to reconstruct realistic recollections for both experience replay training and improvement of the memory module itself. We demonstrate empirically that the module achieves sub-linear scaling with the number of experiences and provides a useful basis for a realistic continual learning system. Our experiments show superior performance over state-of-the-art approaches for lifelong learning with a very small incremental storage footprint.

\section{Related Work} \label{RW}

\textbf{Storing Parameters Instead of Experiences.} Our method is complementary to recent work leveraging episodic storage to stabilize learning \citep{DQN,MFEC,NEC,iCaRL,GEM}. Some recently proposed methods for lifelong learning don't store experiences at all, instead recording the parameters of a network model for each task \citep{PNN,Kirkpatrick17,Pathnet}. This yields linear (or sometimes worse) storage cost scaling with respect to the number of tasks. For our experiments, and in most settings of long-term interest for continual learning, the storage cost of these extra model parameters per task significantly exceeds the per task size of a corresponding experience buffer. 

\textbf{Generative Models to Support Lifelong Learning.} Pseudorehearsals \citep{Robins95} is a related approach for preventing catastrophic forgetting that unlike our recollection module does not require explicit storage of codes. Instead it learns a generative experience model alongside the main model. 
For simple learning problems, very crude approximations of the real data such as randomly generated data from an appropriate distribution can be sufficient. However, for complex problems like those found in NLP and computer vision with highly structured high dimensional inputs, more refined approximations are needed to stimulate the network with relevant old representations. To the best of our knowledge, we are the first to consider variational autoencoders \citep{VAE} as a method of creating pseudo-experiences to support supervised learning. Some recent work \citep{LLGen} considers the problem of generative lifelong learning  for a variational autoencoder, introducing a modified training objective.  This is potentially complementary to our contributions in this paper. 



\textbf{Biological Inspiration and Comparisons.} Interestingly, the idea of scalable experience storage  has a biologically inspired motivation relating back to the pioneering work of \citet{McClelland}, who hypothesized complementary dynamics for the hippocampus and neocortex. In this theory, updated in \citep{CLS}, the hippocampus is responsible for fast learning, providing a very plastic representation for retaining short term memories. Because the neocortex, responsible for reasoning, would otherwise suffer as a result of catastrophic forgetting, the hippocampus also plays a key role in generating approximate \textit{recollections} (experience memories) to interleave with incoming experiences, stabilizing the learning of the neocortex. Our approach follows \textit{hippocampal memory index theory} \citep{HIT86,HIT07}, approximating this role of the hippocampus, with a modern deep neural network model. As this theory suggests, lifelong systems need both a mechanism of \textit{pattern completion} (providing a partial experience as a query for a full stored experience) and \textit{pattern separation} (maintaining separate indexable storage for each experience). As in the theory, we do not literally store previous experiences, but rather compact \textit{indexes} which can be used to retrieve the experience from an \textit{association cortex}, modeled as an auto-encoder.


\textbf{Storing Few Experiences.} Recent work on continual lifelong learning in deep neural networks \citep{iCaRL,GEM} has focused on the resource constrained lifelong learning problem and how to promote stable learning with a relatively small diversity of prior experiences stored in memory. In this work, we complete the picture by also considering the relationship to the \textit{fidelity} of the prior experiences stored memory. We achieve this by considering an additional resource constraint on the number of bits of storage allowed for each experience. 
\section{The Scalable Recollection Module (SRM)} \label{AlgSec} 
\begin{figure}[!h]
    \centering
    \includegraphics[scale=0.4]{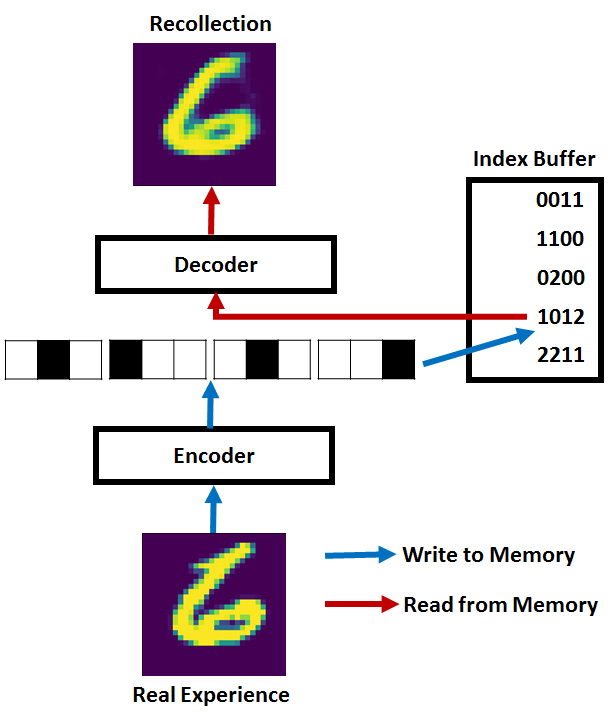}
    \caption{The scalable recollection module.}
    \label{DualModel1}
\end{figure}

The core of our approach is an architecture which supports scalable storage and retrieval of experiences as shown in Figure \ref{DualModel1}.  There are three primary components: an \textit{encoder}, an \textit{index buffer}, and a \textit{decoder}. When a new experience is received (in the figure, an image of the numeral "6"), the encoder compresses it to a  sequence of discrete latent codes (one hot vectors).  These codes are concantenated and further compressed to a $k$ bit binary code or ``index'' shown in decimal in the figure. This compressed code is then stored in the index buffer. This path is shown in blue. Experiences are retrieved from the index buffer by sampling a code from the index buffer and passing it through the decoder to create an approximate reconstruction of the original input. This path is shown in red in the figure.

The recollection buffer is implemented using a discrete variational auto-encoder \citep{CAE1}\citep{CAE2}.  A discrete variational autoencoder is a generative neural model with two components: an encoder and a decoder.  The encoder is trained to take the input (say an image) and produce a discrete distribution over a set of latent categories which describe that input.  To generate data, the discrete distribution is sampled. This can be done in a differentiable way using the so-called ``reparameterization trick" which pushes the (non-differentiable) sampling operation to the front of the network.  The decoder takes the sample and decodes it back into the original form of the input (in our example, an image).  The VAE can then be used to encode experiences into a compact discrete code and later to generate experiences by sampling codes in the latent space and running them through the decoder.

Variational autoencoders have been used in the past to learn generative models of experience (called ``pseudo-rehearsals" in the literature) \citep{Robins95}. Our model differs in two respects: first we are using a discrete VAE which can produce compact codes; and second we are maintaining a buffer of those codes. Typical applications for VAEs focus on a fixed distribution to be learned and require that the VAE reproduce samples from that distribution with high fidelity.  For continual learning it is also important that the VAE adapt to the non-stationary data distribution. Without an index buffer, the VAE's parameters would quickly adapt to the current input and forget its past experiences. Additionally, as we will show later, the buffer leads to greater efficiency in generating samples that capture the variation of the distribution seen.

\section{Improving Experience Replay Training}
The recollection module can be used in many ways in a continual learning setting.  In Algorithm \ref{ER} we show one approach which we will use later in our experiments. 

 
In this setting the model must learn $T$ tasks sequentially from dataset $D$. At every step it receives a triplet $(x,t,y)$ representing the input, task label, and correct output. There are two models to be trained: the recollection module which consists of a memory index buffer $M$, an encoder $ENC_\phi$ and decoder $DEC_\psi$; and a predictive task model $F_\theta$. The training proceeds in two phases: in the first phase we ensure that the recollection module itself is stabilized against forgetting; in the second phase we stabilize the predictive model. 

For the recollection module, we achieve stabilization through a novel extension of experience replay \citep{Lin92}. When an incoming example is received, we first sample multiple batches of recollections from the index buffer and decode them into experiences using the current decoder. We then perform $N$ steps of optimization on the encoder/decoder parameters $\phi$ and $\psi$ by interleaving the current input example with a different batch of past recollections at each of the $N$ optimization steps. For each optimization step, the error for each experience in a batch is computed by encoding that experience into a latent code using the encoder and then decoding back to an experience to compute the reconstruction error. On the first optimization step, the the reconstruction error is computed using the same decoder parameters that were used in the creation of that input experience in the batch.  In subsequent steps, those parameters change as the recollection module is stabilized, learning parameters to successfully reconstruct both the old experiences in the buffer as well as the new experience.  In this way the recollection module continues to \textit{remember} the relevant past experiences in the buffer while integrating new experiences. 

After the recollection module is trained with loss function $\ell_{REC}$ and learning rate $\beta$, the predictive model $F_\theta$ is trained on just one of the recollection sample sets (we arbitrarily chose the first) using loss function $\ell$ and learning rate $\alpha$. Finally, the new sample is written to the index buffer. Perhaps surprisingly, this strategy of reconstructing experiences from codes and then performing experience replay training using them can be as effective for enabling continual learning as replaying real, uncompressed inputs in some cases.



In the resource constrained setting we study in this paper, there is an upper limit on the number of allowable episodic memories $L$ which governs the memory update $M \gets M  \cup \{(z,y)\}$. Our approach is fairly robust to changes in the update rule. For experience replay, we maintain the buffer using reservoir sampling \citep{RS}. In constrast, for the recently proposed Gradient Episodic Memory (GEM) algorithm which modulates gradients on incoming examples by solving a quadratic program with respect to past examples, we follow prior work and keep an equal number of the most recent examples for each task. As detailed in the appendix, integration of Scalable Recollections across algorithms is quite easy and the other differences between the experience replay and GEM algorithms with Scalable Recollections are contained to the different ways of utilizing episodic memory inherent to each lifelong learning algorithm. We note that the PROJECT function within GEM solves a quadratic program explained in \citep{GEM}. 
\begin{algorithm}[h!]
  \caption{Experience Replay Training for Continual Learning with a Scalable Recollection Module}\label{ER}
  \begin{algorithmic}
    \Procedure{Train}{$D,F_\theta,ENC_\phi,DEC_\psi,\alpha,\beta$} 
    \State $M \gets \{\}$
      \For{\texttt{$t = 1,...,T$}}
        \For{\texttt{$x,y$ in $D_t$}}
        
            \State \textbf{Scalable Recollection Module Training:}
          \State \# create $N$ recollection sample sets
            \For{\texttt{s = 1,...,N}}
           \State \# sample latent codes and labels
                \State $z_s,y_s \gets Sample(M) $ 
               \State \# decode the latent codes into recollections  
               \State $x_s \gets DEC_\psi(z_s)$ 
                \State \# save the current label
                \State $Y_s \gets y_s \cup y$
                \State \# save the current recollection  
                \State $X_s \gets x_s \cup x$  
            \EndFor
            \State \# train the recollection module 
            \For{\texttt{$s = 1,...,N$}} 
                \State \# compute recollection module
               gradients 
               \State $u \gets \nabla_{\phi,\psi}\ell_{REC}(DEC_\psi(ENC_\phi(X_s))),X_s)$
               \State \# update the encoder parameters 
                \State $\phi \gets \phi - \beta u_\phi$
                \State \# update the decoder parameters
                \State $\psi \gets \psi - \beta u_\psi$ 
            \EndFor
            
            \State \textbf{Experience Replay Training:}
        \State \# compute main model gradients 
             \State$g \gets \nabla_\theta \ell(F_\theta(X_1,t),Y_1)$ 
            \State \# update the main model parameters \State$\theta \gets \theta - \alpha g$ 
            \State \# encode the recollection sample set \State$z \gets ENC_\phi (x)$ \State \# store it in the index buffer
            \State $M \gets M  \cup \{(z,y)\}$ 
        \EndFor
      \EndFor
      \State \textbf{return} $F_\theta,ENC_\phi,DEC_\psi,M$
    \EndProcedure
  \end{algorithmic}
\end{algorithm}




\section{Recollection Efficiency}

  
In this section we argue that it is more efficient to employ an index buffer rather than sampling directly from the latent VAE code space. This results from the ability of the model to recreate the input distribution. A typical strategy is to randomly sample from old experiences in episodic storage and interleave them with new ones to stabilize learning. So to the extent that the recollections are representative of the full distribution of prior experiences, learning proceeds exactly as if samples were drawn from a stationary i.i.d. distribution.

However, when sampling from the code space of the VAE without a buffer, the sample distribution will be unlikely to match that of the experiences from training.  If the number of examples is larger than the number of possible codes (the capacity of the VAE) then the VAE will be unable to differentiate truly different images and hence have poor reconstruction.  This scenario must be avoided for performance reasons. On the other hand, if the  VAE capacity is considerably larger than the number of training examples, then sampling it at random is highly unlikely to reproduce the input distribution.

To alleviate this problem, a buffer can be added whose distribution of codes will match the distribution of the input data codes, if it is sufficiently large.  Intuitively, setting the buffer size to be less than the capacity of the VAE will ensure that sampling from the buffer is more efficient than sampling directly from the VAE code space.  Our experiments empirically support this hypothesis (Question 6):


 \textbf{Hypothesis 1} \label{Theorem1} \textit{An index buffer is a more efficient parameterization of the input space seen by a variational autoencoder than its latent code if:}
 \[ \ell^c \geq L \]

$\ell$ is the autoencoder latent variable size (i.e. 32 for typical continuous latent variables), $c$ is the number of latent variables in the autoencoders (i.e. the number of hidden units for continuous latent variables), and $L$ is the index buffer size.  

In the vast majority of settings relevant to the study of continual lifelong learning today it is safe to assume that this inequality will hold. For example, even if we store an index for every example in a dataset like MNIST or CIFAR, this inequality will hold unless a continuous latent variable autoencoder has 3 or fewer latent variables. To put it another way, the per sample compression would have to be over 400X for popular datasets like Omniglot and MNIST and well over 1500X for CIFAR. These levels of compression with good reconstruction are far beyond what is possible with any known tools today. Moreover, given, as an example, the blurry nature of CIFAR images, it is possible that this compression quality is completely unfeasible. In practice, do to natural redundancy in the space of samples, it is more likely that $L$ will also only need to be significantly less than the number of examples seen. In our work we find that selecting subsets works well although a buffer may be even more efficient with online clustering strategies as in \citep{kaiser17}. 

\section{Evaluation} \label{LLR} \label{Exp1}

\subsection{Datasets}

Our experiments will primarily focus on three public datasets commonly used for deep lifelong and multi-task learning.

\textbf{MNIST-Rotations:} \citep{GEM} A dataset with 20 tasks including 1,000 training examples for each task. The tasks are random rotations between 0 degrees and 180 degrees of the input space for each digit in MNIST.

\textbf{Incremental CIFAR-100:} \citep{GEM} A continual learning split of the CIFAR-100 image classification dataset considering each of the 20 course grained labels to be a task with 2,500 examples each. 

\textbf{Omniglot:} A character recognition dataset \citep{Omniglot} in which we consider each of the 50 alphabets to be a task. This is an even more challenging setting than explored in prior work on continual lifelong learning, containing more tasks and fewer examples of each class.

\subsection{Evaluation for Continual Lifelong Learning}

In this section we evaluate the benefits of the Scalable Recollection Module in enabling continual lifelong learning. 

\textbf{Metric:} As in prior work, we measure \textit{retention} as our key metric. It is defined as the test set accuracy on all tasks after sequential training has been completed over each task. 

\textbf{Architecture:} We model our experiments after \citep{GEM} and use a Resnet-18 model as $F_\theta$ for CIFAR-100 and Omniglot as well as a two layer MLP with 200 hidden units for MNIST-Rotations. Across all of our experiments, our autoencoder models include three convolutional layers in the encoder and three deconvolutional layers in the decoder. Each convolutional layer has a kernel size of 5. As we vary the size of our categorical latent variable across experiments, we in turn model the number of filters in each convolutional layer to keep the number of hidden variables consistent at all intermediate layers of the network.

\textbf{Module hyperparameters:} In our experiments we used a binary cross entropy loss for both $\ell$ and $\ell_{REC}$. In the appendix we outline a constrained optimization procedure to find the optimal discrete autoencoder latent code design for a given resource footprint constraint. We follow this procedure to derive architectures that can be directly compared to various episodic storage baselines in our experiments. 

\vspace{3mm}
The key question we consider is the following:

\textbf{Question 1} \textit{Is the recollection module useful in improving retention for continual lifelong learning?} 
\vspace{3mm}

To answer this question we compare the retention performance of a system equipped with the recollection module to one equipped with a buffer of real experiences. We consider three datasets: MNIST-Rotations, CIFAR-100, and Omniglot. To account for the fact that real experiences take up more storage, we give both approaches the same storage budget.  Specifically, we define two new quantities: \textit{incremental storage} and \textit{items}. The \textit{incremental storage} is the effective size of the incremental storage used after the sunk cost of the initial model parameters. For clarity we express the effective \textit{incremental storage} size in terms of the number of real examples of storage that would have the same footprint of resources used. The number of \textit{items} by contrast refers to the number of items used in the episodic storage buffer whether they are real or approximate recollections. Obviously, by compressing items stored in the buffer we are able to store more \textit{items} at the same effective \textit{incremental storage} size.

In Table \ref{LooseMNIST} we consider learning with a very small incremental resource footprint on MNIST-Rotations where we allow the effective storage of only one full experience or fewer per class. We find that storing real experiences can perform well in this regime, but the Scalable Recollection Module significantly outperforms them. For example, we achieve considerable improvements over results reported for the popular elastic weight consolidation (EwC) algorithm \citep{Kirkpatrick17} on this benchmark. We note that EwC stores some items for computing the Fisher information. We also note that the large incremental cost of the parameter and Fisher information storage for each task is equal to that of storing a real buffer of over 18 thousand examples.  

In Table \ref{LooseCifar} we show results for Incremental CIFAR-100.  This is a significant challenge for the Scalable Recollection Module since the CIFAR tiny image domain is known to be a particularly difficult setting for VAE performance \citep{TCIFAR1,TCIFAR2}. We explore settings with a very small incremental resource footprint, including a couple of settings with even less incremental memory allowance than the number of classes. Real storage performs relatively well when the number of examples is greater than the number of classes, but otherwise suffers from a biased sampling towards a subset of classes. This can be seen by the decreased performance for small buffer sizes compared to using no buffer at all, learning online. Consistently we see that tuning the recollection module to approximate recollections with a reasonably sized index buffer results in improvements over real storage at the same incremental resource cost. 

To further validate our findings, we tried the Omniglot dataset, attempting to learn continually in the difficult incremental 50 task setting. With an incremental resource constraint of 10 full examples, replay achieves 3.6\% final retention accuracy (online learning produces 3.5\% accuracy). In contrast, the recollection module achieves 5.0\% accuracy. For an incremental resource footprint of 50 full examples, replay achieves 4.3\% accuracy which is further improved to 4.8\% accuracy by taking three gradient descent steps per new example. The recollection module again achieves better performance with 9.3\% accuracy at one step per example and 13.0\% accuracy at three steps per example.

\begin{table}
\small
\centering
\resizebox{0.90\columnwidth}{!}{
\begin{tabular}{|l|c|c|c|}
\hline
Method & Incremental Storage & Items & Retention \\ \hline
GEM Real Storage & 100 & 100 & 62.5 \\
 & 	200 & 200 & 67.4 \\ \hline
GEM Recollections & 100 & 3000 & \textbf{79.0} \\
 & 200 & 3000 & \textbf{81.5} \\ \hline
Replay Real Storage & 100 & 100	& 63.4 \\
 & 200 & 200 & 71.3 \\ \hline
Replay Recollections & 100 & 3000 & \textbf{75.6} \\
 & 200 & 3000 & \textbf{81.1} \\ \hline
EwC & 18288 & 1000 & 54.6 \\ \hline
Online & 0 & 0 & 51.9 \\ \hline
\end{tabular}
}
\caption{\label{LooseMNIST} \small Retention results on MNIST-Rotations for low effective buffer sizes with an incremental storage resource constraint. }
\end{table}

\begin{table}
\centering
\resizebox{0.9\columnwidth}{!}{
\begin{tabular}{|l|c|c|c|}
\hline
Model & Incremental Storage & Items & Retention \\ \hline
Online & 0 & 0 & 33.3 \\
LwF \citep{LwF} & 0 & 0 & 34.5 \\ \hline 
Replay Real Storage & 10 & 10 & 29.4 \\ 
	& 50 & 50 & 33.4 \\ 
	& 200 & 200	& 43.0 \\ \hline
Replay Recollections	& 10 & 5000	& \textbf{39.7} \\ 
	& 50	& 5000	& \textbf{47.9} \\ 
	& 200	& 5000	& \textbf{51.6} \\ \hline
GEM Real Storage & 20 & 20 & 23.4 \\ 
	& 60 & 60 & 40.6 \\ 
	& 200 & 200	& 48.7\\ \hline
GEM Recollections	& 20 & 5000	& \textbf{52.4} \\ 
	& 60	& 5000	& \textbf{56.0} \\ 
	& 200	& 5000	& \textbf{59.0} \\ \hline
\end{tabular}
}
\caption{\label{LooseCifar} \small Incremental CIFAR-100 results for low effective incremental buffer sizes. GEM requires sizes that are multiples of $T=20$. }
\end{table}

\vspace{3mm}
\textbf{Question 2} \textit{How does use of Scalable Recollections influence the long term retention of knowledge?}
\vspace{3mm}

\begin{figure}[!b]
    \centering
    \includegraphics[scale=0.17]{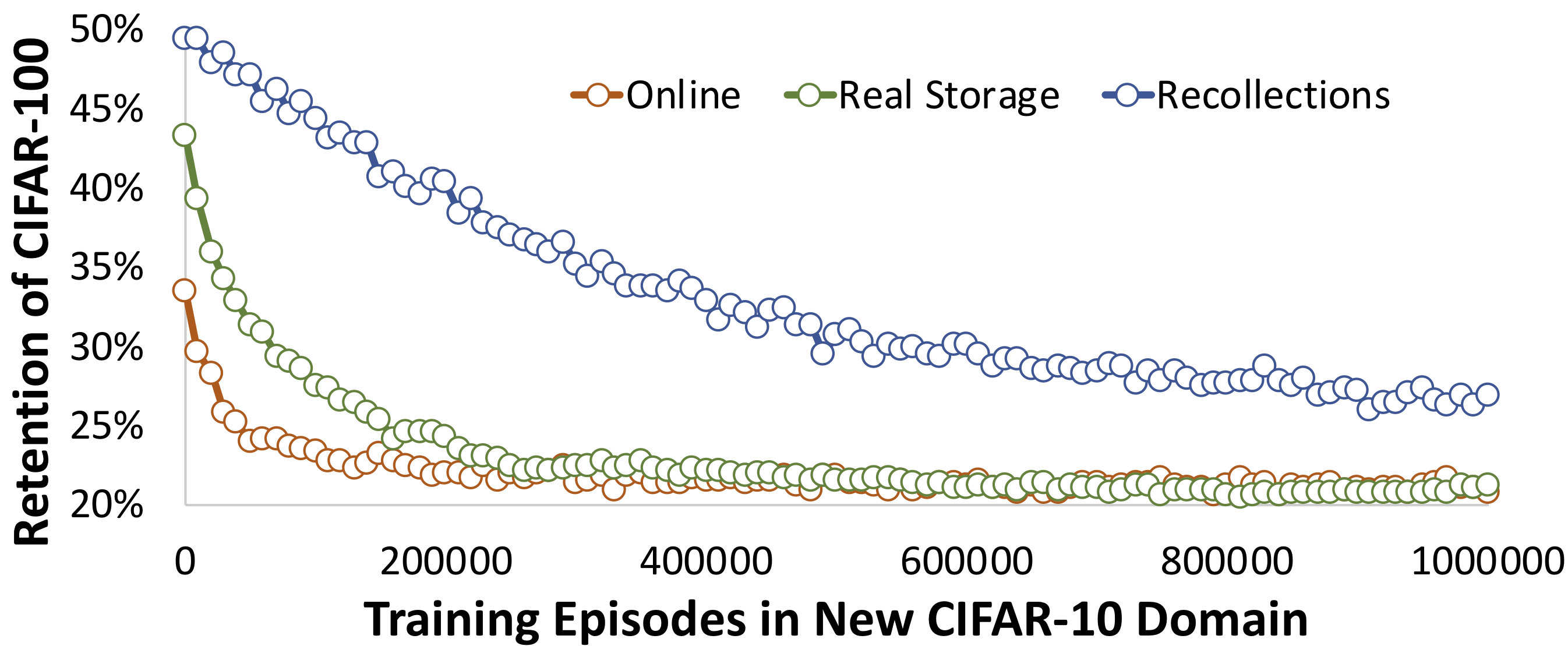}
    \caption{Retained accuracy on CIFAR-100 after prolonged training on CIFAR-10. CIFAR-10 contains images with a similar structure, but is drawn from a disjoint set of labels. }
    \label{SimpleRetention}
\end{figure}

The value of using recollections becomes even more apparent for long term retention of skills. We demonstrate this in Figure \ref{SimpleRetention} by first training models on Incremental CIFAR-100 and then training them for 1 million training examples on CIFAR-10. The number of training examples seen from CIFAR-100 is only 5\% of the examples seen from CIFAR-10. Not only does the recollection module allow experience replay to generalize more effectively than real storage during initial learning, it also retains the knowledge much more gracefully over time. We provide a detailed chart in the appendix 
that includes learning for larger real storage buffer sizes as a comparison. A six times larger real storage buffer loses knowledge much faster than Scalable Recollections despite better performance during training on CIFAR-100. 

\vspace{3mm}
\textbf{Question 3} \textit{Can Scalable Recollections overcome the overhead of autoencoder model parameters?}
\vspace{3mm}

To achieve the greater goals of lifelong learning, we are mostly interested in scaling to conditions where the number of examples seen is very large and as a result the incremental storage footprint dominates the overhead of the initial model parameters. However, we would like to empirically demonstrate that we can overcome this overhead in practical problems. Unfortunately, to demonstrate performance with a very small total storage footprint on a single dataset, an incredibly small autoencoder would then be required to learn a function for the very complex input space from scratch. We demonstrate in Table \ref{TransferExp} that transfer learning provides a solution to this problem. By employing unlabeled background knowledge we are able to perform much better at the onset with a small autoencoder. We explore a total resource footprint on top of the size of $F_\theta$ equivalent to 200 real examples. This equates to the smallest setting explored in \citep{GEM} and we demonstrate that we are able to achieve state of the art results by initializing only the autoencoder representation with one learned on CIFAR-10. CIFAR-10 is drawn from the same larger database as CIFAR-100, but is non-overlapping. 

\begin{table}

\centering
\resizebox{0.9\columnwidth}{!}{
\begin{tabular}{|l|c|c|}
\hline
Model & Items & Retention   \\ \hline
Replay Real Storage & 200 & 43.0    \\ \hline
Replay Recollections - No Transfer & 1392 & 43.7   \\
Replay Recollections - CIFAR-10 Transfer & 1392 & 49.7   \\ \hline
GEM Real Storage & 200 & 48.7    \\ \hline
GEM Recollections - No Transfer & 1392 & 43.7   \\
GEM Recollections - CIFAR-10 Transfer & 1392 & \textbf{54.2}   \\ \hline
iCaRL (Rebuffi et al., 2017) & 200 & 43.6    \\ \hline
\end{tabular}}
\caption{\label{TransferExp}\small Retention results on Incremental CIFAR-100 with a 200 real episode effective total storage resource footprint.}
\end{table}

\subsection{Why Scalable Recollections Work}

Given the impressive performance of the Scalable Recollections Module for supporting the continual lifelong learning for neural networks, we would like to further explore the proposed system to elucidate why it works so well.  

\vspace{3mm}
\textbf{Question 4} \textit{How do discrete latent codes compare with continuous latent codes for compression?}
\vspace{3mm}

\begin{figure}[!h]
    \centering
    \includegraphics[scale=0.21]{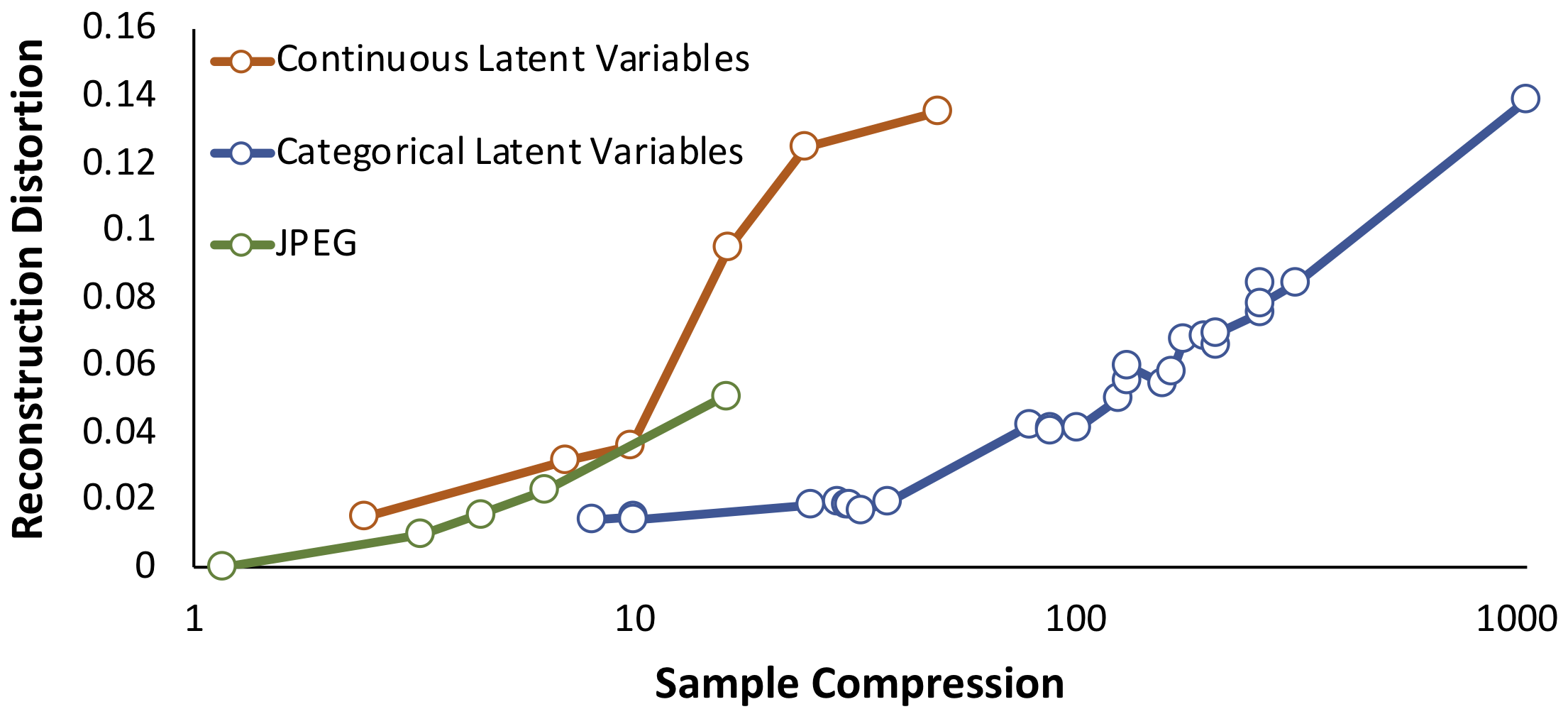}
    \caption{Comparing reconstruction L1 distance on the MNIST training set and sample compression for continuous latent variable and categorical latent variable autoencoders.}
    \label{DistvComp}
\end{figure}

In Figure \ref{DistvComp} we empirically demonstrate that autoencoders with categorical latent variables can achieve significantly more storage compression of input observations at the same average distortion as autoencoders with continuous variables. In this experiment to make the continuous baseline even tougher to beat on the training set, we leverage a standard autoencoder instead of a variational one as it does not add noise to its representation, which would make it harder to reconstruct the original input. See the appendix for details. 

\vspace{3mm}
\textbf{Question 5} \textit{How do learned methods of compression compare with static compression algorithms?}
\vspace{3mm}

In Figure \ref{DistvComp} we also compare the performance of autoencoders with JPEG, which is a static compression algorithm commonly used in industry. We can see that JPEG performs quite well for low degrees of compression, but scales less gracefully than discrete autoencoder based compression for larger degrees of sample compression. This is because learned compression algorithms have the ability to further customize to the regularities seen in the data than a generic one. More detail is provided in the appendix.

\vspace{3mm}
\textbf{Question 6} \textit{Do we see gains in learning efficiency as a result of the index buffer as predicted by Hypothesis 1?}
\vspace{3mm}

The recollection module must not only provide a means of compressing the storage of experiences in a scalable way, but also a mechanism for efficiently sampling recollections so that they are truly representative of prior experiences. 

\begin{table*}
\small
\centering
\resizebox{1.6\columnwidth}{!}{
\begin{tabular}{|c|c|c|c|}
\hline
Latent Representation & Sampling Strategy & Reconstruction Distortion & Nearest Neighbor Distortion \\ \hline
38 2d variables & Code Sampling & 0.058 & 0.074 \\ 
 & Buffer Sampling & 0.058 & 0.054 \\ \hline
168 2d variables & Code Sampling & 0.021 & 0.081 \\
 & Buffer Sampling & 0.021 & 0.021 \\ \hline
\end{tabular}}
\caption{Comparing the nearest training example L1 distance of code and buffer sampling based recollections averaged across 10,000 samples. Reconstruction distortion of the autoencoder is measured on the test set and is not influenced by the strategy.} \label{Tradeoff}
\end{table*}

\begin{figure}[!h]
    \centering
    \includegraphics[scale=0.25]{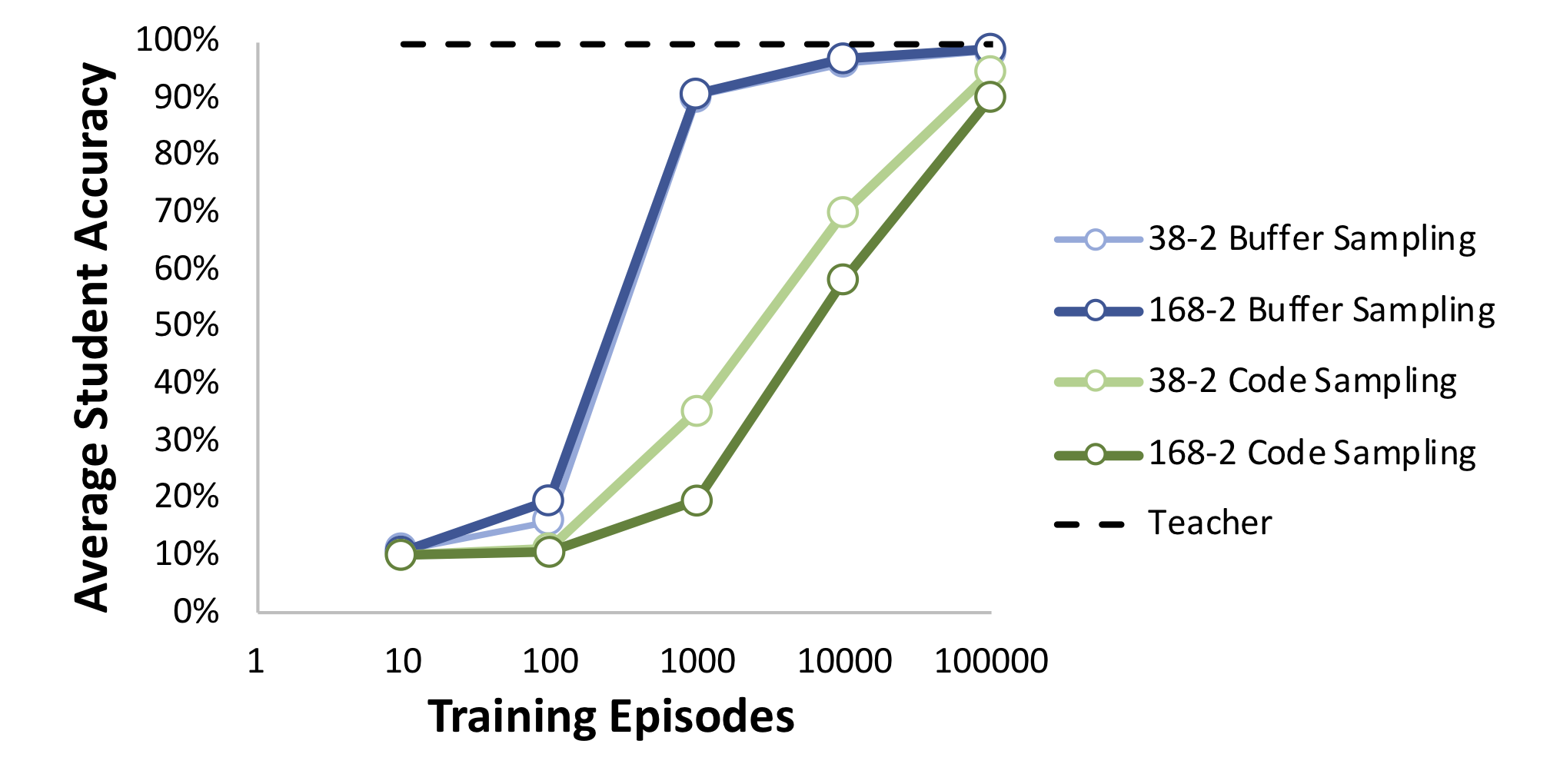}
    \caption{Comparison of generative transfer learning performance using a CNN teacher and student model on MNIST while using code sampling and recollection module sampling.}
    \label{Episodes} 
\end{figure}

We first consider the typical method of sampling a variational autoencoder, which we will refer to as \textit{code sampling}, where each latent variable is selected randomly. Obviously, by increasing the capacity of the autoencoder we are able to achieve lower reconstruction distortion. However, interestingly, we find that while increasing the autoencoder capacity increases modeling power, it also increases the chance that a randomly sampled latent code will not be representative of those seen in the training distribution. Instead, we maintain an index buffer of indexes associated with prior experiences. Let us call sampling from the index buffer, \textit{buffer sampling}.  Table \ref{Tradeoff} shows a comparison of code and buffer sampling for two different latent variable representation sizes.  The reconstruction distortion is the error in reconstructing the recollection using the decoder.  The nearest neighbor distortion is the distance from the sampled code to its nearest neighbor in the training set. We can see that for the same reconstruction distortion, the buffer approach yields a significantly smaller nearest neighbor distortion.  This means that the buffer sampling produces a more representative sample than code sampling.

How much does this matter in practice? Figure \ref{Episodes} demonstrates its utility through a knowledge distillation experiment.  Here we compare the two representation sizes and approaches at the task of distilling a CNN teacher model into a student model of the same architecture through the latent codes.  That is the student is trained on the reconstructed data using the teacher output as a label. By far the best learning curve is obtained using buffer sampling. We would like to emphasize that these results are not a byproduct of increased model capacity associated with the buffer: the small representation with the buffer significantly outperforms the big representation with code sampling despite 7.4x fewer total bits of storage including the model parameters and buffer. In the appendix we include comprehensive experiments showing that distillation based on buffer sampling with a discrete latent code VAE is even more efficient than storing real examples.

\vspace{1mm}
\textbf{Question 7} \textit{Does recollection based self-stabilization of the autoencoder lead to effective continual lifelong learning?}
\vspace{-2mm}

\begin{figure}[!b]
    \centering
    \includegraphics[scale=0.21]{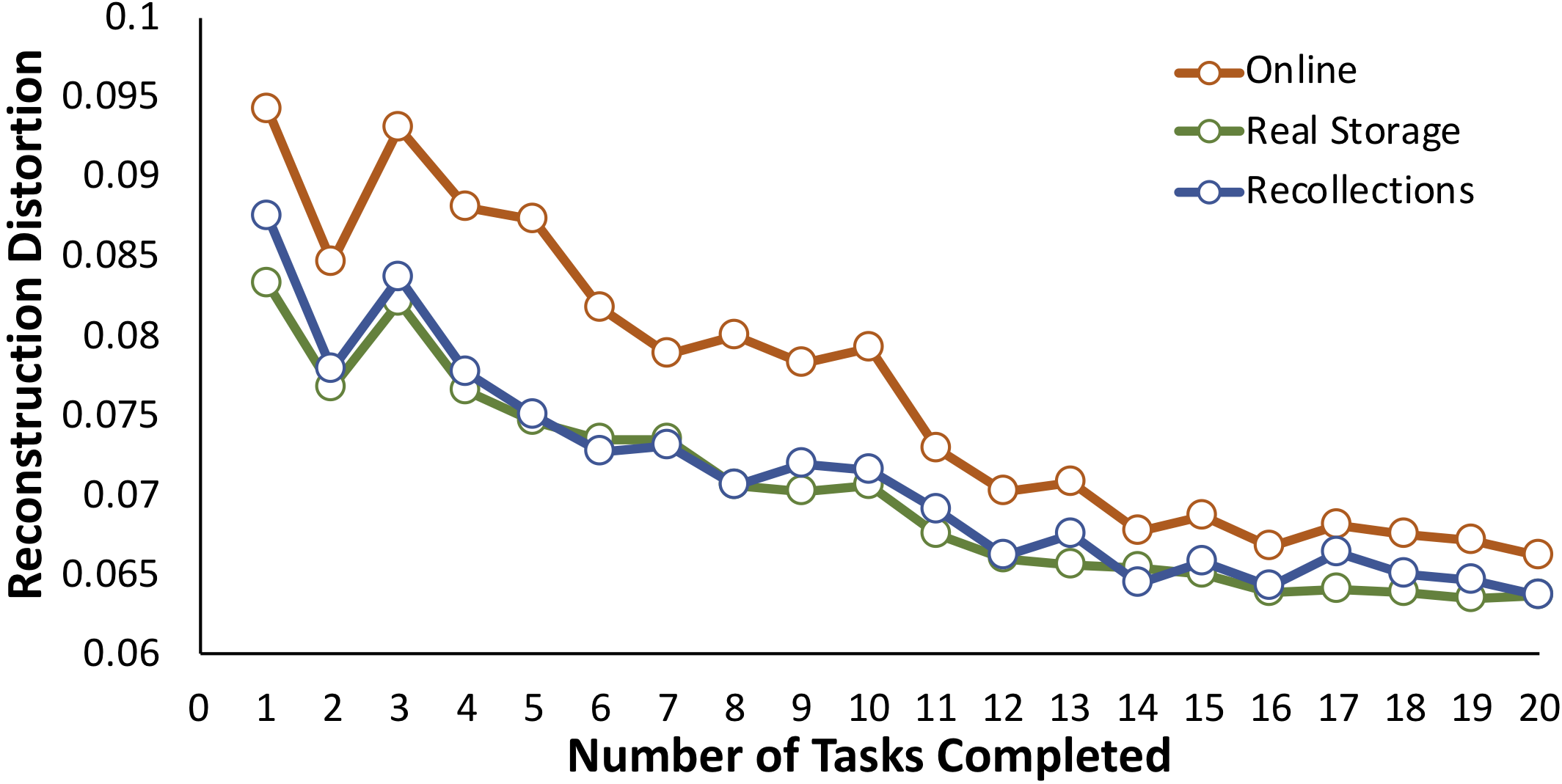}
    \caption{Test set L1 reconstruction distortion on Incremental CIFAR-100 with an effective incremental buffer size of 200.}
    \label{SelfGen} 
\end{figure}

We validate our recollection module training procedure by demonstrating that recollections generated by an autoencoder model can actually be effective in preventing catastrophic forgetting for the very same model. 
As shown in Figure \ref{SelfGen} for continual learning on CIFAR-100 with number of steps $N=10$ and an effective incremental buffer size of an average of two items per class, the recollection module is very similarly effective to real storage for stabilizing the lifelong autoencoder. The negative effects of the less effective synthetic examples are apparently drowned out by the positive effects of a larger diversity of stored examples.

\vspace{1mm}
\textbf{Question 8} \textit{Can recollection efficiency improve over time? }
\vspace{-2mm}

\begin{figure}[!b]
    \centering
    \includegraphics[scale=0.24]{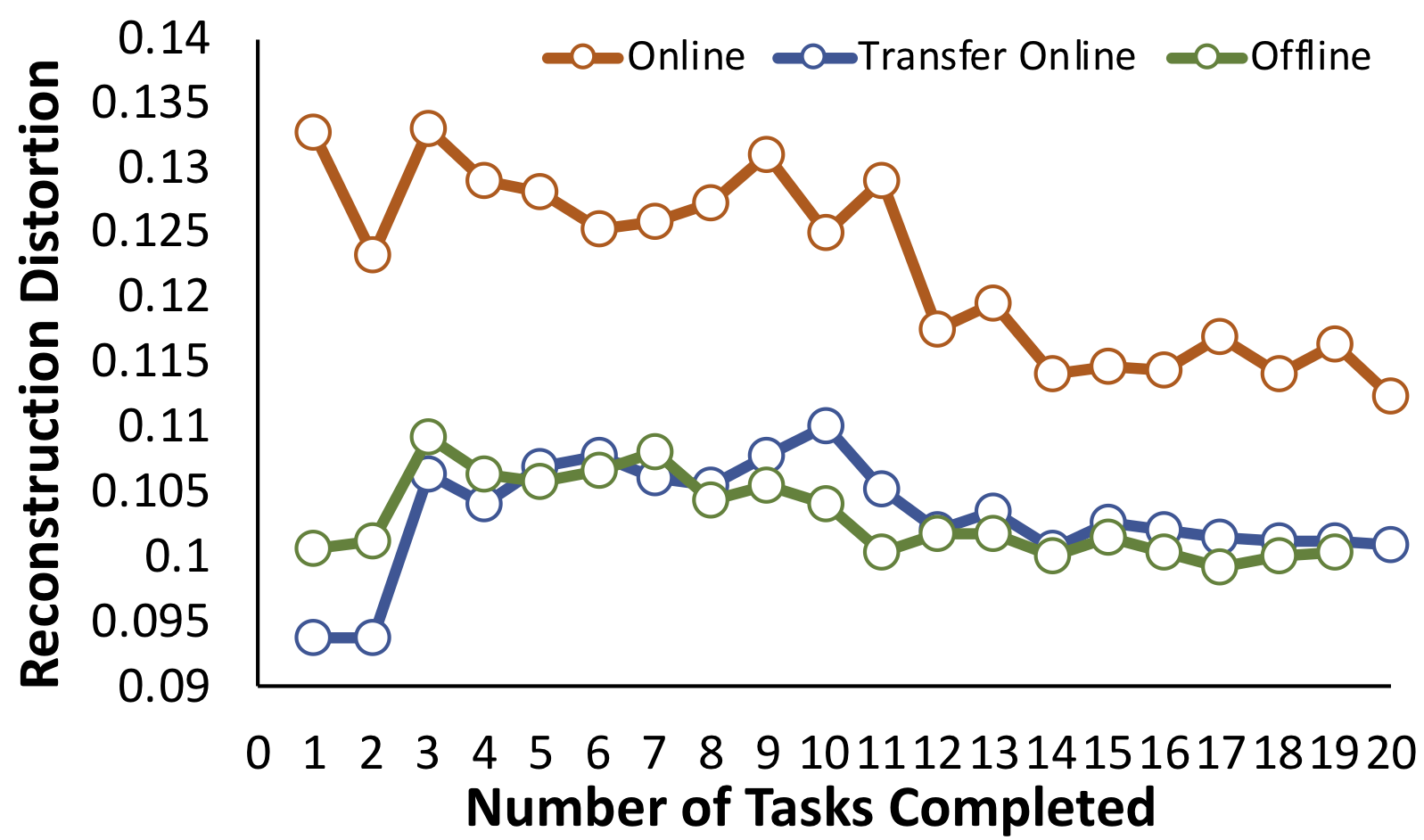}
    \caption{Test set L1 reconstruction distortion on Incremental CIFAR-100 of a 76 2d categorical latent variable autoencoder.}
    \label{TransferAE}
\end{figure}

We explore this question in Figure \ref{TransferAE} where we consider online training of the model with a random initialization and no buffer (Online) and offline training with random initialization and full data storage (Offline) trained over 100 iterations.  Predictably, access to unlimited storage and all of the tasks simultaneously means that the performance of Offline is consistently better than Online. To demonstrate the value of transfer from a good representation in the continual learning setting, we additionally plot an online model with no replay buffer and a representation initialized after training for 100 iterations on CIFAR-10 (Transfer Online).  Transfer adds significant value, performing comparably to the randomly initialized model with access to all tasks simultaneously and unlimited storage (Offline). In fact, it performs considerably better for the first few tasks where the number of prior experiences is much greater than the number of new experiences. Improvements from transfer learning  thus have a substantial effect in stabilizing $F_\theta$ as well as demonstrated in Table \ref{TransferExp}. 

\section{Conclusion} \label{Conclusion}

We have proposed and experimentally validated a general purpose Scalable Recollection Module that is designed to scale for very long time-frames. We have demonstrated superior performance over other state-of-the-art approaches for lifelong learning using very small incremental storage footprints. These increases can be dramatically boosted with unsupervised recollection module pre-training. 
We have shown that VAEs with categorical latent variables significantly outperform those with continuous latent variables (and even JPEG) for lossy compression. Finally, we have also shown that maintaining an explicit buffer is key to capturing the distribution of previously seen samples and generating realistic recollections needed to effectively prevent forgetting. 





\bibliographystyle{aaai} 
\bibliography{iclr2018_conference}

\appendix

\section{Recollection Module Implementation} \label{Meth1}
The key role of the scalable recollections module is to efficiently facilitate the transfer of knowledge between two neural models. In this section we argue that the recently proposed discrete latent variable variational autoencoders \citep{CAE1,CAE2} are ideally suited to implement the encoder/decoder functionality of the recollection module. 

Typical experience storage strategies store full experiences which can be expensive.  For example, to store 32x32 CIFAR images with 3 color channels and 8-bits per pixel per channel will incur a cost per image stored of 8x3x32x32 = 24,576 bits. Deep non-linear autoencoders are a natural choice for compression problems. An autoencoder with a continuous latent variable of size $h$, assuming standard 32-bit representations used in modern GPU hardware, will have a storage cost of 32$h$ bits for each latent representation. Unfortunately, continuous variable autoencoders which use 32-bits for their network parameters may incur an unnecessary storage cost for many problems -- especially in constrained-resource settings.

A solution which combines the benefit of VAE training with an ability to explicitly control precision is the recently proposed VAE with categorical latent variables \citep{CAE1,CAE2}. Here we consider a bottleneck representation between the encoder and decoder with $c$ categorical latent variables each containing $l$ dimensions representing a \emph{one hot} encoding of the categorical variable. This can be compressed to a binary representation of $k = c \cdot \lceil \log_2(l) \rceil$ bits.

\subsection{Discrete VAE Implementation}

In order to model an autoencoder with discrete latent variables, we follow the success of recent work \citep{CAE1,CAE2} and employ the Gumbel-Softmax function. The Gumbel-Softmax function leverages the Gumbel-Max trick \citep{gumbel,gumbel2} which provides an efficient way to draw samples $z$ from a categorical distribution with class probabilities $p_i$ representing the output of the encoder: 

\begin{equation}
z = \mathrm{one\_hot}(argmax_i[g_i+\log(p_i)])
\label{eq:gumbelmax}
\end{equation}

In equation \ref{eq:gumbelmax}, $g_i$ is a sample drawn from Gumbel(0,1), which is calculated by drawing $u_i$ from Uniform(0,1) and computing $g_i$=-log(-log($u_i$)). The $one\_hot$ function quantizes its input into a one hot vector. The softmax function is used as a differentiable approximation to argmax, and we generate $d$-dimensional sample vectors $y$ with temperature $\tau$ in which: 

\begin{equation}
y_i = \frac{\exp((g_i+\log(p_i))/\tau)}{\sum_{j=1}^{d} \exp((g_j+\log(p_j))/\tau)}
\label{eq:gumbelsoftmax}
\end{equation}

The  Gumbel-Softmax  distribution  is  smooth  for $\tau$ $>$ 0,  and  therefore  has  a  well-defined  gradient with respect to the parameters $p$. During forward propagation of the categorical autoencoder, we send the output of the encoder through the sampling procedure of equation \ref{eq:gumbelmax} to create a categorical variable. However, during backpropagation we replace non-differentiable categorical samples with a differentiable approximation during training using the Gumbel-Softmax estimator in equation \ref{eq:gumbelsoftmax}. Although past work \citep{CAE1,CAE2} has found value in varying $\tau$ over training, we still were able to get strong results keeping $\tau$ fixed at 1.0 across our experiments. Across all of our experiments, our generator model includes three convolutional layers in the encoder and three deconvolutional layers in the decoder.

One issue when deploying an autoencoder is the determination of the latent code size hyperparameter. In the next section
we will derive a maximization procedure that we use throughout our experiments to find the optimal autoencoder latent variable size to use for a given resource constraint.

\subsection{Details on GEM Integration} \label{GEMInt}
In this subsection we outline the training procedure for Gradient Episodic Memory (GEM) \citep{GEM} integrated with our proposed scalable recollection module in Algorithm \ref{GEM}.

\begin{algorithm}
  \caption{GEM Training for Continual Learning with a Scalable Recollection Module}\label{GEM}
  \begin{algorithmic}
    \Procedure{Train}{$D,F_\theta,ENC_\phi,DEC_\psi,\alpha,\beta$} 
    \State $M \gets \{\}$
      \For{\texttt{$t = 1,...,T$}}
        \For{\texttt{$(x,y)$ in $D_t$}}
            \State \# Scalable Recollection Module Training:
            \For{\texttt{$s = 1,...,N$}}
                \State $z_s,y_s \gets Sample(M) $
                \State $x_s \gets DEC_\psi(z_s)$
                \State $Y_s \gets y_s \cup y$
                \State $X_s \gets x_s \cup x$ 
            \EndFor
            \For{\texttt{$s = 1,...,N$}}
                \State $u \gets \nabla_{\phi,\psi}\ell_{REC}(DEC_\psi(ENC_\phi (X_s))),X_s)$

                \State $\phi \gets \phi - \beta u_\phi$
                \State $\psi \gets \psi- \beta u_\psi$
            \EndFor
            \State \# GEM Training:
            \State $z \gets ENC_\phi (x)$
            \State $M_t \gets M_t \cup (z,y)$
            \State $g \gets \nabla_\theta \ell(F_\theta(x,t),y)$
            \State $g_k \gets \nabla_\theta \ell(F_\theta,DEC_\psi(M_k))$ for all $k < t$
            \State $\overline{g} \gets \Call{PROJECT}{g,g_{1},...,g_{t-1}}$
            \State $\theta \gets \theta - \alpha \overline{g}$
        \EndFor
      \EndFor
      \State \textbf{return} $F_\theta,ENC_\phi,DEC_\psi,M$
    \EndProcedure
  \end{algorithmic}
\end{algorithm}

\section{Optimizing Latent Code Size for a Resource Constraint} \label{LEC}

The ability of a discrete variational autoencoder to memorize inputs should be strongly related to the effective bottleneck capacity $C_\mathrm{ve}$, which we define, for discrete latent variables, as:     

\begin{equation}
C_\mathrm{ve} = \log_2 l^c\, .
\label{eq:capacity}
\end{equation}

\subsection{Incremental Storage Resource Constraints} \label{loose}

First, let us consider the dynamics of balancing resources in a simple setting where we have an incremental storage constraint for new incoming data without regard for the size of the model used to compress and decompress recollections. We refer to the total storage constraint over all $N$ incoming examples as $\gamma$ and the average storage rate limit as $\gamma/N$. We can then define $\rho$ as the probability that an incoming example is stored in memory. Thus, the expected number of bits required per example stored is $\rho S_\mathrm{sbe}$, assuming simple binary encoding. If we treat $\rho$ as fixed, we can then define the following optimization procedure to search for a combination of $c$ and $l$ that maximizes capacity while fulfilling an incremental resource storage constraint:

\begin{equation}
\begin{aligned}
& \underset{c,l}{\text{maximize}}
& & C_\mathrm{ve} \\
& \text{subject to}
& & \rho S_\mathrm{sbe} \leq \frac{\gamma}{N}\, , \
\end{aligned}
\label{eq:tradeoff1}
\end{equation}

which yields the approximate solution $C_\mathrm{ve} \simeq \frac{\gamma}{N\rho}$. As seen in equation \ref{eq:tradeoff1}, there is an inherent trade-off between between the diversity of experiences we store governed by $\rho$ and the distortion achieved that is related to the capacity. The optimal trade-off is likely problem dependent. Our work takes a first step at trying to understand this relationship. For example, we demonstrate that deep neural networks can see improved stabilization in resource constrained settings by allowing for some degree of distortion. This is because of an increased ability to capture the diversity in the data at the same incremental resource constraint.  

\subsection{Total Storage Resource Constraints} \label{strict}

In some ways, the incremental storage constraint setting described in the previous section is not the most rigorous setting when comparing recollections to a selected subset of full inputs. Another important factor is the number of parameters in the model $|\theta|$ used for compression and decompression. $|\theta|$ generally is also to some degree a function of $c$ and $l$. For example, in most of our experiments, we use the same number of hidden units $cl$ at each layer as used in the bottleneck layer. With fully connected layers, this yields $|\theta|(c,l) \propto (cl)^2$. As such, we can revise equation \ref{eq:tradeoff1} to handle a more rigorous constraint for optimizing a discrete latent variable autoencoder architecture:

\begin{equation}
\begin{aligned}
& \underset{c,l}{\text{maximize}}
& & C_\mathrm{ve} \\
& \text{subject to}
& & \rho S_\mathrm{sbe} + |\theta|(c,l) \leq \gamma/N, \
\end{aligned}
\label{eq:tradeoff2}
\end{equation}

While this setting is more rigorous when comparing to lossless inputs, it is a somewhat harsh restriction with which to measure lifelong learning systems. This is because it is assumed that the compression model's parameters should be largely transferable across tasks. To some degree, these parameters can be viewed as a sunk cost from the standpoint of continual learning. In our experiments, we also look at transferring these representations from related tasks to build a greater understanding of this trade-off.   

\section{Additional Details on Experimental Protocol}
\label{AppendixA}

Each convolutional layer has a kernel size of 5. As we vary the size of our categorical latent variable across experiments, we in turn model the number of filters in each convolutional layer to keep the number of hidden variables consistent at all intermediate layers of the network. In practice, this implies that the number of filters in each layer is equal to $cl/4$. We note that the discrete autoencoder is stochastic, not deterministic and we just report one stochastic pass through the data for each experimental trial. In all of our knowledge distillation experiments, we report an average result over 5 runs. For MNIST and Omniglot we follow prior work and consider 28x28 images with 1 channel and 8-bits per pixel. MNIST and Omniglot images were originally larger, but others have found the down sampling to 28x28 does not effect performance of models using it to learn.

\subsection{Distortion as a Function of Compression Experiments}
\label{AppendixA1}

More detail about the architecture used in these experiments are provided for categorical latent variables in Table \ref{cllossy} and for continuous latent variables in Table \ref{cllossy2}. For each architecture we ran with a learning rate of 1e-2, 1e-3, 1e-4, and 1e-5, reporting the option that achieves the best training distortion. For the distortion, the pixels are normalized by dividing by 255.0 and we take the mean over the vector of the absolute value of the reconstruction to real sample difference and then report the mean over the samples in the training set. Compression is the ratio between the size of an 8bpp MNIST image and the size of the latent variables, assuming 32 bits floating point numbers in the continuous case and the binary representation for the categorical variables. The JPEG data points were collected using the Pillow Python package using quality 1, 25, 50, 75, and 100. We subtracted the header size form the JPEG size so it is a relatively fair accounting of the compression for a large data set of images all of the same size. The JPEG compression is computed as an average over the first 10,000 MNIST training images.

\begin{table}
\small
\centering
\begin{tabular}{|c|c|c|c|}
\hline
$c$  & $l$ & Compression & Distortion \\ \hline 
6 & 20 & 209.067 & 0.06609 \\ \hline
10 & 20 & 125.440 & 0.04965 \\ \hline
6 & 16 & 261.333 & 0.07546 \\ \hline
12 & 10 & 130.667 & 0.05497 \\ \hline
10 & 14 & 156.800 & 0.05410 \\ \hline
24 & 3 & 130.667 & 0.05988 \\ \hline
38 & 2 & 165.053 & 0.05785 \\ \hline
6 & 2 & 1045.333 & 0.13831 \\ \hline
40 & 3 & 78.400 & 0.04158 \\ \hline
20 & 2 & 313.600 & 0.08446 \\ \hline
8 & 6 & 261.333 & 0.08423 \\ \hline
12 & 6 & 174.222 & 0.06756 \\ \hline
30 & 2 & 209.067 & 0.06958 \\ \hline
24 & 6 & 87.111 & 0.04065 \\ \hline
4 & 37 & 261.333 & 0.07795 \\ \hline
8 & 15 & 196.000 & 0.06812 \\ \hline
48 & 10 & 32.667 & 0.01649 \\ \hline
209 & 8 & 10.003 & 0.01455 \\ \hline
12 & 37 & 87.111 & 0.03996 \\ \hline
313 & 4 & 10.019 & 0.01420 \\ \hline
392 & 3 & 8.000 & 0.01348 \\ \hline
50 & 18 & 25.088 & 0.01859 \\ \hline
168 & 2 & 37.333 & 0.01955 \\ \hline
108 & 3 & 29.037 & 0.01894 \\ \hline
62 & 2 & 101.161 & 0.04073 \\ \hline
208 & 2 & 30.154 & 0.01832 \\ \hline
68 & 5 & 30.745 & 0.01849 \\ \hline
\end{tabular}
\caption{\label{cllossy} This table provide more specifics about the discrete latent variable architectures involved in Figure 2 of the main text.}
\end{table}

\begin{table}
\small
\centering
\begin{tabular}{|c|c|c|c|}
\hline
$h$ & Compression & Distortion \\ \hline 
1 & 49 & 0.135196 \\ \hline 
2 & 24.5 & 0.124725 \\ \hline 
3 & 16.33333333 & 0.0947032 \\ \hline 
5 & 9.8 & 0.0354035 \\ \hline 
7 & 7 & 0.031808 \\ \hline 
20 & 2.45 & 0.0149272 \\ \hline 
\end{tabular}
\caption{\label{cllossy2} This table provide more specifics about the continuous latent variable architectures involved in 
Figure 2 of the main text.} 
\end{table}

\subsection{Continual Lifelong Learning Experiments}
\label{AppendixA5}

In our experiments on MNIST-Rotations we found it optimal to set the GEM learning rate to 0.1 and memory strength to 0.5. In our recollections experiments, we used an autoencoder with 104 4d variables for the size 100 buffer and one with 139 8d variables for the size 200 buffer. For experience replay, we used a learning rate of 0.01 for the predictive model and a learning rate of 0.001 for the autoencoder. For the size 100 buffer we experienced more over-fitting to the buffer and used a batch size of 5. For the size 200 buffer we used a batch size of 25.

Our categorical latent variable autoencoders had the following sizes for Incremental CIFAR-100: 48 2d variables for an effective buffer size of 10, 98 2d variables for an effective buffer size of 20, 244 2d variables for an effective buffer size of 50, 294 2d variables for an effective buffer size of 60, and 620 3d variables for an effective buffer size of 200. The predictive model was trained with a learning rate of 1e-3 in all of our replay experiments and 0.1 in all of our GEM experiments. The learning rate for the autoencoder was 1e-4 for the buffer size of 200 regardless of the lifelong learning model and buffer size of 20 with GEM. The autoencoder learning rate was 1e-3 for replay with buffer sizes 10 and 50 as well as GEM with buffer size 60. The GEM memory strength parameter was set to 0.1 for our real storage experiments and 0.01 for our experiments with recollection. In our experiments with GEM and with replay on MNIST-Rotations, we were able to get very good results even just training our autoencoder online and not leveraging the recollection buffer to stabilize its training. We could have seen further improvements, as we show on CIFAR-100, by using the buffer for stabilization.

For incremental Omniglot our learning rate was set to 1e-3. For the effective buffer size of 50 experiments, we leveraged a categorical latent variable autoencoder with 312 2d variables. For the effective buffer size of 10 experiments, we utilized a categorical latent variables consisting of 62 2d variables. We follow 90\% multi-task training and 10\% testing splits for Omniglot established in \citep{OmniMTL}.

During the transfer learning experiments from CIFAR-10, for replay a learning rate of 1e-3 was used for the Resnet-18 reasoning model and a learning rate of 3e-4 was used for the discrete autoencoder generator. For GEM, we used a learning rate of 0.1 for the resnet model, a learning rate of 1e-3 for the autoencoder, and a memory strength of 0.1. For the experiment without transfer learning, we instead used a higher learning rate of 1e-3 for the replay autoencoder. We used a learning rate of 1e-4 for the autoencoder in our transfer learning GEM experiments.  

\subsection{Detailed Retention Results} \label{AppRet}
\begin{figure}[!b]
    \centering
    \includegraphics[scale=0.7]{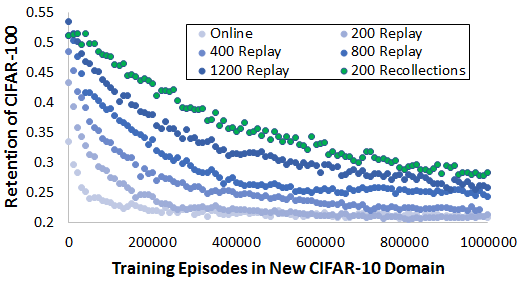}
    \caption{Retention of performance on CIFAR-100 after prolonged training on CIFAR-10. We compare recollections and full storage replay buffer strategies listed by their effective incremental buffer size. }
    \label{Retention}
    \vspace{-4mm}
\end{figure}

We provide a detailed version of Figure 2 
in the main text
in Figure \ref{Retention}. It includes learning for larger real storage buffer sizes as a comparison. For example, a six times larger real storage buffer looses knowledge significantly faster than scalable recollections despite better performance when originally training on Incremental CIFAR-100. 

\section{Scalable Recollections for Distillation}

\vspace{3mm}
\textbf{Question 9} \textit{Can Scalable Recollections retain knowledge so that it is transferable to new neural network instantiations?}
\vspace{3mm}

\begin{table*}
\small
\centering
\resizebox{1.7\columnwidth}{!}{
\begin{tabular}{|c|c|c|c|c|c|c|c|c|}
\hline \bf Episodes & \bf Real & \bf 10\% & \bf  2\% & \bf 1\% & \textbf{Real $x$} & \bf 10x & \bf 50x & \bf 100x  \\
 & \bf Data & \bf Sample & \bf Sample & \bf Sample & \textbf{ Teacher $y$} & \bf Compress & \bf Compress & \bf Compress \\ \hline
10 & 10.43 & 9.94 & 11.07 & 10.70 & 10.07 & 10.65 & 10.99 & \textbf{13.89}\\ \hline 
100 & 19.63 & 18.16 & 22.82 & 22.35 & \textbf{25.32} & 19.34 & 16.20 & 21.06 \\ \hline 
1000 & 90.45 & 88.88 & 90.71 & 89.93 & \textbf{91.01} & 90.66 & 90.52 & 90.03 \\ \hline 
10000 & 97.11 & 96.83 & 95.98 & 94.97 & \textbf{97.42} & 96.77 & 96.37 & 95.65 \\ \hline 
100000 & 98.51 & 97.99 & 96.14 & 94.92 & \textbf{98.63} & 98.59 & 98.17 & 97.75 \\ \hline 
\end{tabular}
}
\caption{\label{RandomCNN} \small Knowledge distillation from a CNN teacher model to a randomly initialized student model of the same architecture on MNIST as a function of the number of episodes used to train the student model and the method that the teacher model uses to preserve episodic storage from the training set. }
\end{table*}

In our experiments, we train a teacher model with a LeNet \citep{LeNet} convolutional neural network (CNN) architecture on the popular MNIST benchmark, achieving 99.29\% accuracy on the test set. We would like to test whether recollections drawn from our proposed recollection module are sufficient input representations for the teacher neural network to convey its function to a seperate randomly initialized student neural network of the same size. In Table \ref{RandomCNN} we validate the effectiveness of our technique by comparing it to some episodic storage baselines of interest. As baselines we consider training with the the same number of randomly sampled real examples, using real input and the teacher's output vector as a target, and using random sampling to select a subset of real examples to store. When training with a large number of memories for a more complete knowledge transfer, the recollection compression clearly shows dividens over random sampling baselines. This is impressive particularly because these results are for the stricter total storage footprint compression setting where we account for the autoencoder model parameters and on a per sample basis the compression is actually 37x, 101x, and 165x.  

We also would like to validate these findings in a more complex setting for which we consider distillation with outputs from a 50 task Resnet-18 teacher model that gets 94.86\% accuracy on Omniglot. We test performance after one million training episodes, which is enough to achieve teacher performance using all of the real training examples. However, sampling diversity restricts learning significantly, for example, achieving 28.87\% accuracy with 10\% sampling, 8.88\% with 2\% sampling, and  5.99\% with 1\% sampling. In contrast the recollection module is much more effective, achieving 87.86\% accuracy for 10x total resource compression, 74.03\% accuracy for 50x compression, and 51.45\% for 100x compression. Later we also demonstrate that these results generalize to mismatched architectures for the student and teacher. Moreover, in the we will also show that by using heuristics to explore the buffer instead of random sampling we can transfer knowledge even more efficiently to a new model than with randomly drawn real examples. Indeed, it is clear that Scalable Recollections provide a general purpose reservoir of knowledge that is efficiently transferable to anything ranging from old network instantiations of the same model (in which case it reinforces prior learning) to totally new and different architectures (looking to obtain this knowledge from scratch). 

\subsection{MNIST Generative Distillation Experiments}
\label{AppendixA3}

Alongside the teacher model, we train an variational autoencoder model with discrete latent variables. Each model is trained for 500 epochs. During the final pass through the data, we forward propogate through each training example and store the latent code in an index buffer. This buffer eventually grows to a size of 50,000. After training is complete, the index buffer is used as a statistical basis for sampling diverse recollections to train a student network. A logical and effective strategy for training a student model is to sample randomly from this buffer and thus capture the full distribution. For all of our distillation experiments we ran the setting with a learning rate of 1e-3 and 1e-4, reporting the best result. We found that the higher learning rate was beneficial in setting with a low number of examples and the lower learning rate was beneficial in setting with a larger number of examples.  The categorical latent variable autoencoders explored had the following representation sizes: 168 2d variables for 10x compression, 62 2d variables for 50x compression, and 38 2d variables for 100x compression. For our code sampling baselines, we used the numpy random integer function to generate each discrete latent variable. 

\subsection{Omniglot Generative Distillation Experiment}
\label{AppendixA4}

The learning rate for the Resnet-18 reasoning model was 1e-4 in our experiments. Our trained discrete autoencoder models were of the following representation sizes: 32 variables of size 2 for 100x compression, 50 variables of size 2 for 50x compression, and 134 variables of size 2 for 10x compression. We follow 90\% multi-task training and 10\% testing splits for Omniglot established in \citep{OmniMTL}.

\subsection{CNN to MLP Distillation Results}
In Figure \ref{RandomMLP} we explore generative knowledge distillation transferring knowledge from a CNN teacher network to a MLP student network. 

\label{AppendixB}
\begin{table*}
\small
\centering
\begin{tabular}{|c|c|c|c|c|c|c|c|c|}
\hline \bf Episodes & \bf Real & \bf 10\% & \bf  2\% & \bf 1\% & \textbf{Real $x$} & \bf 10x & \bf 50x & \bf 100x  \\
 & \bf Data & \bf Sample & \bf Sample & \bf Sample & \textbf{ Teacher $y$} & \bf Compress & \bf Compress & \bf Compress \\ \hline
10 & 13.64 & \textbf{17.04} & 14.57 & 15.13 & 15.87 & 16.70 & 11.80 & 14.66  \\ \hline
100 & 36.37 & 37.04 & 38.35 & 34.04 & 38.56 & 37.16 & 40.09 & \textbf{42.31} \\ \hline
1000 & 80.54 & 79.08 & 78.18 & 77.76 & 80.00 & \textbf{80.72} & 80.00 & 77.75 \\ \hline
10000 & 91.04 & 90.84 & 88.38 & 86.83 & 90.86 & \textbf{91.37} & 90.60 & 90.46  \\ \hline
100000 & 96.66 & 95.02 & 91.61 & 88.97 & 96.60 & \textbf{96.71} & 96.24 & 95.22  \\ \hline
\end{tabular}
\caption{\label{RandomMLP} Generative knowledge distillation random sampling experiments with a CNN teacher and MLP student model on MNIST. }
\end{table*}

\subsection{Automated Generative Curriculum Learning} \label{AAD}

\begin{table*}
\small
\centering
\begin{tabular}{|c|c|c|c|c|c|c|c|c|}
\hline \bf Episodes & \bf Real & \bf \textbf{Real $x$} & \bf Active 10x & \bf Active 100x & \bf Active \& Diverse & \bf Active \& Diverse  \\
 & \bf Data & \textbf{ Teacher $y$} & \bf Compress & \bf Compress & \bf 10x Compress & \bf 100x Compress \\ \hline
10 & 10.43 & 10.07 & 9.95 & 10.19 & 10.67 & \textbf{11.51} \\ \hline
100 & 19.63 & 25.32 & 14.80  & 22.57 & 27.05 & \textbf{29.93} \\ \hline
1000 & 90.45 & 91.01 & 93.45 & 92.97 & \textbf{94.81} & 92.54 \\ \hline
10000 & 97.11 & 97.42 & \textbf{98.61} & 97.53 & 98.59 & 97.66\\ \hline
100000 & 98.51 & 98.63 & 99.18 & 98.25 & \textbf{99.20} & 98.32 \\  \hline
\end{tabular}
\caption{\label{ActiveCNN} Generative knowledge distillation active and diverse sampling experiments with a CNN teacher and student model on MNIST. The real input baselines are randomly sampled.}
\end{table*}

While random sampling from a buffer can be very effective, we would like to further maximize the efficiency of distilling knowledge from a teacher model to a student model. This motivates the automated curriculum learning setting \citep{Bengio2009} as recently explored for multi-task learning in \citep{Graves17} or rather automated generative curriculum learning in our case. We tried some simple reinforcement learning solutions with rewards based on \citep{Graves17} but were unsuccessful in our initial experiments because of the difficulty of navigating a complex continuous action space. We also tried an active learning formulation proposed for GANs to learn the best latent code to sample \citep{BC} at a given time. We had limited success with this strategy as well as it tends to learn to emphasize regions of the latent space that optimize incorrectness, but no longer capture the distribution of inputs. 

\textbf{Designing generative sampling heuristics.} Inspired by these findings, we instead employ simple sampling heuristics to try to design a curriculum with prototypical qualities like responsiveness to the student and depth of coverage. We model responsiveness to the student as \textit{active sampling} by focusing on examples where the student does not have good performance. We randomly sample $k$ latent codes using our recollection buffer and choose the one that is most difficult for the current student for backpropagation by cheaply forward propagating through the student for each. By sampling from the recollection buffer, we are able to ensure our chosen difficult samples are still representative of the training distribution. We set $k$ to 10 in our experiments so the sampling roughly equates to sampling once from the most difficult class for the student model at each point in time. We model depth of coverage by sampling a bigger batch of random examples and adding a filtering step before considering difficulty. We would like to perform \textit{diverse sampling} that promotes subset diversity when we filter from $kn$ examples down to $k$ examples. One approach to achieving this is a Determinantal Point Process (DPP) \citep{DPP} as recently proposed for selecting diverse neural network mini-batches \citep{DPPNN}. We use the dot product of the inputs as a measure of similarity between recollections and found the DPP to achieve effective performance as a diverse sampling step. However, we follow \citep{MSSS} and use a process for sampling based on the sum of the squared similarity matrix as outlined in the next section. We found the sum of the squared similarity matrix to be equally effective to the determinant and significantly more scalable to large matrices. We also set $n$ to 10 in our experiments. 

\subsection{Minimum Sum of Squared Similarities}
\label{MSSS}
This algorithm is trying to find a new landmark point that maximizes the determinant by finding a point that minimizes the sum of squared similarities (MSSS). The MSSS algorithm initially randomly chooses two points from the dataset $X$. It then computes the sum of similarities between the sampled points and a subset, $T$, selected randomly from the remaining data points. The point with the smallest sum of squared similarities is then picked as the next landmark data point. The procedure is repeated until a total of $m$ landmark points are picked.

\begin{algorithm}[H]
   \caption{The Minimum Sum of Squared Similarities Algorithm}
\label{alg:MSSS} 
\begin{algorithmic}[1]
 \State {\bfseries }\textbf{Input:} 
   $X=\{x_1,x_2,...,x_n\}$: dataset \\
   $m$: number of landmark data points\\
   $\gamma$: size of the subsampled set from the remaining data, in percentage \\ 
 \State {\bfseries }\textbf{Output:} $\widetilde{S} \in R^{m \times m}$: similarity matrix between landmark points
 \State {\bfseries }   Initialize $\widetilde{S}=I_0$
 \State {\bfseries }   \textbf{For (i$=$0 to i$<$2) do}
 \State {\bfseries }   \hspace{2em} $\widetilde{x}_i=Random(X)$
 \State {\bfseries }   \hspace{2em} $\widetilde{S}:=\widetilde{S}_{\cup x_i} $
  \State {\bfseries }   \hspace{2em} $\widetilde{X}:=\widetilde{X} \cup \{\widetilde{x}_i\} $
 \State {\bfseries }  \textbf{End For} 
  \State {\bfseries } \textbf{While $i<m$ do}
   \State {\bfseries } \hspace{2em} $T=Random(X\backslash \{\widetilde{X}\},\gamma)$
  \State {\bfseries } \hspace{2em} Find $\widetilde{x}_{i}=argmin_{x \in T} \sum_{j<i-1} sim^2(x,\widetilde{x}_j)$
  \State {\bfseries }  \hspace{2em} $\widetilde{S}:=\widetilde{S}_{\cup \widetilde{x}_i} $
   \State {\bfseries }   \hspace{2em} $\widetilde{X}:=\widetilde{X} \cup \{\widetilde{x}_i\} $
 \State {\bfseries }\textbf{End While}     
   \end{algorithmic}
\end{algorithm}

\end{document}